\documentclass{article}

\usepackage[final, nonatbib]{nips_2017}

\usepackage[utf8]{inputenc} 
\usepackage[T1]{fontenc}    
\usepackage{hyperref}       
\usepackage{url}            
\usepackage{booktabs}       
\usepackage{amsfonts}       
\usepackage{nicefrac}       
\usepackage{microtype}      

\usepackage[square,numbers]{natbib}

\usepackage{amsmath}
\usepackage{amsfonts}
\usepackage{amssymb}
\usepackage{amsthm}

\usepackage{graphicx}
\usepackage{caption}
\usepackage{subcaption}
\usepackage{wrapfig, booktabs}

\newcommand{\mb}[1]{\mathbf{#1}}

\title{Deep Learning with Permutation-invariant Operator for Multi-instance Histopathology Classification}

\author{
  Jakub M. Tomczak\\
  University of Amsterdam\\
  \And
  Maximilian Ilse\\
  University of Amsterdam\\
  \And
  Max Welling\\
  University of Amsterdam\\
}

\begin{document}

\maketitle

\begin{abstract}
The computer-aided analysis of medical scans is a longstanding goal in the medical imaging field. Currently, deep learning has became a dominant methodology for supporting pathologists and radiologist. Deep learning algorithms have been successfully applied to digital pathology and radiology, nevertheless, there are still practical issues that prevent these tools to be widely used in practice. The main obstacles are low number of available cases and large size of images (a.k.a. the \textit{small n, large p} problem in machine learning), and a very limited access to annotation at a pixel level that can lead to severe overfitting and large computational requirements. We propose to handle these issues by introducing a framework that processes a medical image as a collection of small patches using a single, shared neural network. The final diagnosis is provided by combining scores of individual patches using a permutation-invariant operator (combination). In machine learning community such approach is called a multi-instance learning (MIL).
\end{abstract}

\section{Introduction}

Deep learning has become a leading tool for analyzing medical images, and digital pathology as its major application area \citep{LKBSCGLGS:17}. Main practical issues in current deep learning methods for medical imaging are low number of recorded cases, large size of images (slides) and low availability of a diagnosis with a pixel level annotation (a.k.a. \textit{weakly labeled data}). These problems lead to severe overfitting, impractical computations, \textit{e.g.}, training using images larger than $250\times 250$ pixels requires already a considerably large amount of computational resources, and difficulties in information flow from single label for large images. We propose to handle these issues by introducing a framework that processes a medical image as a collection of small patches using a single, shared neural network. The final diagnosis is provided by combining scores of individual patches. In machine learning community such approach is called a \textit{multi-instance learning} (MIL) \citep{ML:98}.

\paragraph{Related work} There are different approaches to MIL with various combining operators \citep{CS:14, KRL:91, RD:00, RKBDR:08, ZPV:06} but these methods were mainly used for already pre-processed data. Recently, there is an increase of interest in applying MIL to medical imaging and, especially, to histopathology. One of first such methods used SVM and Boosting to cluster and classify colon cancer images \citep{XZCT:12}. Recently, a single neural network with a MIL-pooling layer was used to  classify and segment microscopy images with populations of cells \citep{KBF:16}. A method that is closely related to our approach utilized a neural network to process small patches in the first stage of training and the Expectation Maximization algorithm to determine latent labels of the patches in the second stage \citep{HSKGDS:16}. However, our model is trained end-to-end by backpropagation.

\section{Methodology}

\paragraph{Problem statement} A classical supervised learning problem aims at finding a model that takes an object, $\mb{x} \in \mathbb{R}^{D}$, and predicts a value of a target variable, $y \in \{0,1\}$. In the multi-instance learning problem, however, there is a bag of objects, $\mathcal{X}_{K} = \{\mb{x}_{1}, \ldots, \mb{x}_{K}\}$, that exhibit neither dependency nor ordering among each other. There is also a single label associated with this bag. We assume that $K$ could vary for different bags. We do not have access to individual labels of the objects within the bag, \textit{i.e.}, we assume $y_{1}, \ldots, y_{K}$ are unknown, but we know that the label of the bag is $1$ if at least one object is $1$, \textit{i.e.}, $y = 1 \iff \exists_k : y_{k}=1$. This statement is equivalent to the logic OR operator and could be further re-formulated as the maximum operator: $y = \max_{k} \{ y_{k} \}.$ The max-operator is \textbf{permutation-invariant} that is an important property since objects within a bag are independent. 

Training a bag-level classifier requires a permutation-invariant combination of individual labels $y_{k}$ that are given by an instance-level (shared) classifier. In this paper, we propose to train a model using the likelihood approach. We take the Bernoulli distribution for the bag label:
\begin{equation}\label{eq:bernoulli}
p(y|\mathcal{X}_{K}) = \big{(} \theta(\mathcal{X}_{K}) \big{)}^{y}\ \big{(}1 - \theta(\mathcal{X}_{K}) \big{)}^{1-y},
\end{equation}
where $\theta(\mathcal{X}_{K}) \in [0,1]$ is the probability of $y=1$ given the bag of objects $\mathcal{X}_{K}$. Further, we consider a shared instance-level classifier (a neural network) with parameters $\psi$, $f_{\psi}(\mb{x}_{k})$, that returns a score for the $k$-th object, $z_k = f_{\psi}(\mb{x}_{k})$ and $z_k \in [0,1]$. Then, the parameter $\theta(\mathcal{X}_{K})$ is modeled using a permutation-invariant operator $g: [0,1]^{K} \rightarrow [0,1]$, \textit{i.e.}, $\theta(\mathcal{X}_{K}) = g \big{(} f_{\psi}(\mb{x}_{1}), \ldots, f_{\psi}(\mb{x}_{K}) \big{)}$.

\paragraph{Permutation-invariant operators} Obviously, we can choose the max-operator as $g$ but it is not necessarily well-suited for training neural networks using the backpropagation. Alternatively, we consider the following differentiable operators:
\begin{itemize}
\item[\textbf{(i)}] Noisy-Or (NOR) operator \citep{HS:13}:
$$\theta(\mathcal{X}_{K}) = 1 - \prod_{k=1}^{K} \big{(} 1 - f_{\psi}(\mb{x}_{k}) \big{)},$$
\item[\textbf{(ii)}] Integrated Segmentation and Recognition (ISR) operator \citep{KRL:91}: 
$$\theta(\mathcal{X}_{K}) = \frac{ \sum_{k=1}^{K} v_{k} }{ 1 + \sum_{k=1}^{K} v_{k} },$$
where $v_{k} = \frac{ f_{\psi}(\mb{x}_{k}) }{ 1 - f_{\psi}(\mb{x}_{k}) }$,
\item[\textbf{(iii)}] Log-sum-exp (LSE) operator \citep{RD:00} with $r > 0$:
$$\theta(\mathcal{X}_{K}) = \frac{1}{r} \ln \frac{1}{K} \sum_{k=1}^{K} \exp\big{(} r f_{\psi}(\mb{x}_{k}) \big{)}.$$
\end{itemize}

\paragraph{Training} Once the operator is chosen, we train the model by minimazing the negative log-likelihood using (\ref{eq:bernoulli}):
\begin{equation}
\mathcal{L}(\psi) = -\frac{1}{N} \sum_{n} \ln p(y_{n}|\mathcal{X}_{K,n}, \psi) .
\end{equation}

\section{Workflow}

\begin{figure}[!hbtp]
\begin{minipage}[t]{1.0\linewidth}
  \centering
  \centerline{\includegraphics[width=\textwidth]{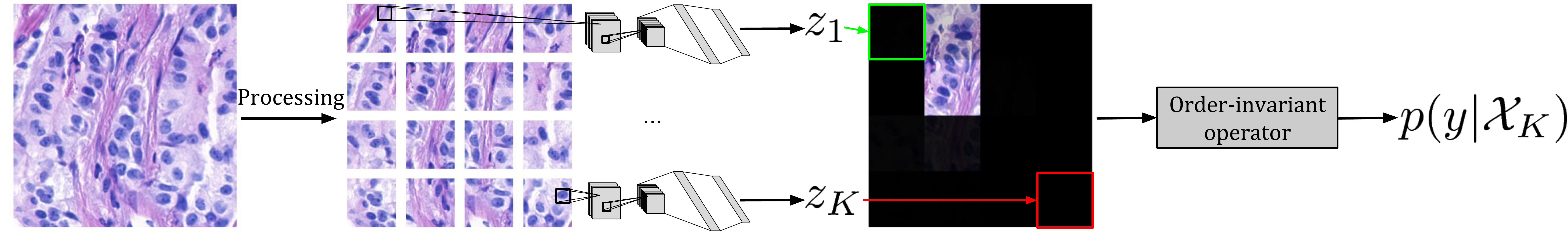}}
  \caption{A schematic representation of the proposed workflow.}
  \label{fig:architecture}
\end{minipage}
\end{figure}

In our framework the input is a slide or a patch from a needle biopsy stained with Hematoxylin \& Eosin (H\&E). Further, we divide the input into small patches (\textit{e.g.}, $96\times 96$ pixels). Each small patch is processed by a shared neural network $f_{\psi}(\mb{x}_{k})$, which consists of several convolutional layers and fully-connected layers with dropout, and it returns a score of each small patch, $z_{k}$. A larger score determines a Region of Interest (ROI) that could be later presented to a human doctor. Eventually, an application of a permutation-invariant operator provides the probability of a diagnosis, \textit{e.g.}, benign or malignant tumor. The proposed framework is presented in Figure \ref{fig:architecture}.

\section{Experiments}

\paragraph{Data} In the experiments we used a dataset that consists of 58 H\&E stained histopathology image excerpts (896$\times$768 pixels) taken from 32 benign and 26 malignant breast cancer patients \citep{GBOM:08}. Due to a limited size of the dataset, a 4-fold cross-validation is used as in \citep{KZH:14}. For images in the training set, we select eight 768$\times$768 overlapping subimages. However, for images in the test set we select a single 768$\times$768 subimage from the center of the image. During training, we use 10\% of the training set for validation and monitoring a training progress. Subsequently, each subimage is divided into patches of 96$\times$96 pixels. A patch is discarded if more than 75\% of the pixels are white.

\begin{figure}[!hbpt]
	\centering
	\includegraphics[width=.9\linewidth]{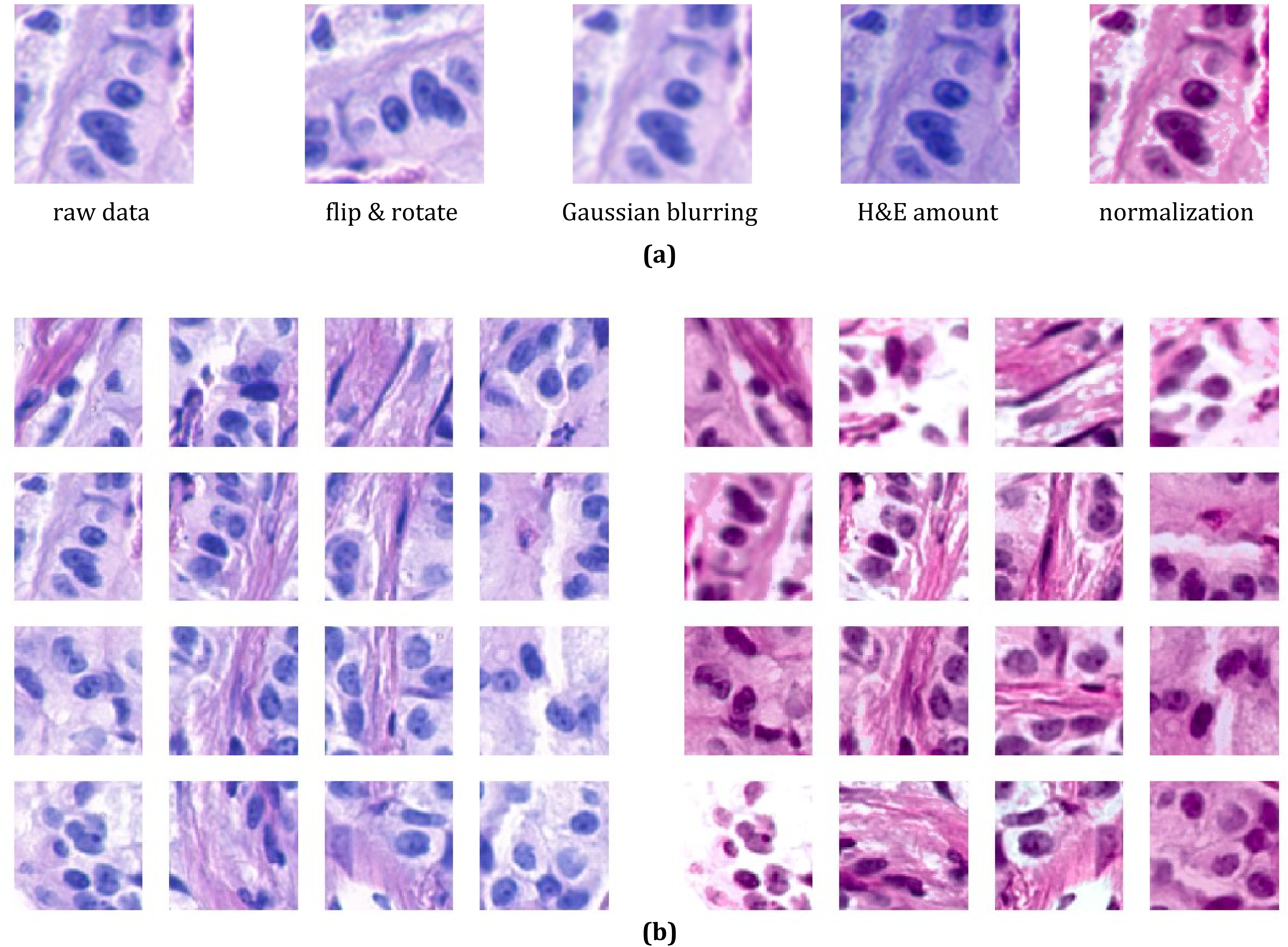}
	\caption{\textbf{(a)} Four different data augmentation transformations applied to a raw data. \textbf{(b)} An image divided into patches (on the left) processed by data augmentation transformations (on the right).}
	\label{fig:data_augmentation}
\end{figure}

\paragraph{Data augmentation} In every training iteration we perform data augmentation to prevent overfitting. We randomly adjust the amount of H\&E by decomposing the RGB color of the tissue into the H\&E color space \citep{RJ:01}, followed by multiplying the magnitude of H\&E for a pixel by two i.i.d. Gaussian random variables with expectation equal to one. We randomly rotate and mirror every patch. Lastly, we blur the patch using a Gaussian blur filter with a randomly chosen blur radius. See Figure \ref{fig:data_augmentation} for examples of data augmentation transformations.

\paragraph{Results and discussion} We compared our approach (\textsc{Deep\{NOR,ISR,LSE\}-MIL}) with the Gaussian process multi-instance learning (\textsc{GPMIL}) and its relational extension \textsc{RGPMIL} \citep{KZH:14}. Results are given in Table \ref{tab:results}.

First, we notice that the proposed approach achieved similar performance to Gaussian process-based methods in terms of AUC. Second, the LSE operator failed to obtain high accuracy and F-score but it still resulted in high AUC. Comparing all operators, we believe that Noisy-or is the most promising but in order to obtain even better results a kind of regularization is required. A possible extension of the presented work would be an application of the Bayesian learning similarly to \citep{RKBDR:08}. However, we leave investigating these issues for further research. 

\begin{table}[!hbpt]
  \centering
  \caption{Results of the 4-cross-validation on the breast cancer data.}
  \vspace{0mm}
  \begin{tabular}{cccccc}
    \textsc{Method}						& \textsc{Accuracy} & \textsc{Precision} & \textsc{Recall} & \textsc{F-score} & \textsc{AUC} \\
    \midrule
    \textsc{GPMIL} \citep{KZH:14}		& N/A				& N/A				 & N/A			   & N/A			  & 0.86 \\
    \textsc{RGPMIL} \citep{KZH:14}		& N/A				& N/A				 & N/A			   & N/A			  & \textbf{0.90} \\    
    \textsc{DeepNOR-MIL}				& \textbf{0.879}	& 0.828				 & \textbf{0.923}  & \textbf{0.873}	  & 0.88 \\ 
    \textsc{DeepISR-MIL}				& 0.828				& 0.808				 & 0.808		   & 0.808			  & \textbf{0.90} \\
    \textsc{DeepLSE-MIL} ($r=10$)		& 0.621				& \textbf{0.833}	 & 0.192		   & 0.312			  & 0.88 \\
    \bottomrule
  \end{tabular}
  \label{tab:results}
\end{table}

\subsubsection*{Acknowledgments}
Jakub M. Tomczak was funded by the European Commission within the Marie Sk\l odowska-Curie Individual Fellowship (Grant No. 702666, ''Deep Learning and Bayesian Inference for Medical Imaging''). Maximilian Ilse was funded by the Nederlandse Organisatie voor Wetenschappelijk Onderzoek (Grant ''DLMedIa: Deep Learning for Medical Image Analysis'').


\end{document}